\documentclass[]{interact}

\usepackage{epstopdf}
\usepackage{subfigure}

\usepackage{natbib}
\bibpunct[, ]{(}{)}{,}{a}{}{,}
\renewcommand\bibfont{\fontsize{10}{12}\selectfont}

\theoremstyle{plain}

\theoremstyle{definition}

\theoremstyle{remark}

\usepackage{color}

\begin{document}

\articletype{Journal}

\title{Deploying machine learning to assist digital humanitarians:  making image annotation in OpenStreetMap more efficient}

\author{\name{John E. Vargas-Mu\~{n}oz\textsuperscript{a*}\thanks{*First author. Email: john.vargas@ic.unicamp.br}, Devis Tuia\textsuperscript{b}, Alexandre X. Falc\~{a}o\textsuperscript{a}}
\affil{\textsuperscript{a} Laboratory of Image Data Science, Institute of Computing,\\ University of Campinas, Campinas, Brazil;\\\textsuperscript{b} Laboratory of Geo-information Science and Remote Sensing,\\ Wageningen University \& Research, the Netherlands}
}

\maketitle

\begin{abstract}
\textbf{This is a preprint of the paper published in the International Journal of Geographical Information Science. The publisher version can be found here:}\\ \textbf{https://doi.org/10.1080/13658816.2020.1814303}\\
Locating populations in rural areas of developing countries has attracted the attention of humanitarian mapping projects since it is important to plan actions that affect vulnerable areas. Recent efforts have tackled this problem as the detection of buildings in aerial images. However, the quality and the amount of rural building annotated data in open mapping services like OpenStreetMap (OSM) is not sufficient for training accurate models for such detection. Although these methods have the potential of aiding in the update of rural building information, they are not accurate enough to automatically update the rural building maps. In this paper, we explore a human-computer interaction approach and propose an interactive method to support and optimize the work of volunteers in OSM. The user is asked to verify/correct the annotation of selected tiles during several iterations and therefore improving the model with the new annotated data. The experimental results, with simulated and real user annotation corrections, show that the proposed method greatly reduces the amount of data that the volunteers of OSM need to verify/correct. The proposed methodology could benefit humanitarian mapping projects, not only by making more efficient the process of annotation but also by improving the engagement of volunteers.
\end{abstract}


\section{Introduction}
\label{sec:introduction}

A large amount of mapping information of buildings has been collected in open and commercial mapping services like OpenStreetMap and Google maps. 
However, the building mapping data is concentrated mainly in urban areas.
Recently, humanitarian organizations like, the Humanitarian OSM Team (HOT) and the Red Cross have created projects to map buildings in rural areas of developing countries, in order that Non-Governmental Organizations (NGOs) can use the maps to plan actions in response to crises that affect those areas. For instance, the HOT Task manager project~\footnote{https://tasks.hotosm.org/} provides a platform that includes a web-based map editor in which volunteers manually annotate buildings and other classes of interest in some predefined areas in OSM. The volunteers verify/annotate all the tiles in a selected geographical area by observing aerial imagery. This is a time-consuming task since most of the tiles do not contain the objects of interest in rural areas.  

\begin{figure}[!t]
\begin{center}
\begin{tabular}{ccc} 
  \includegraphics[width=0.25\columnwidth]{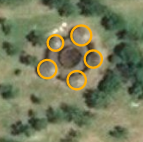} & 
  \includegraphics[width=0.25\columnwidth]{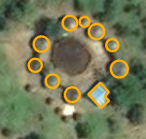} &
  \includegraphics[width=0.25\columnwidth]{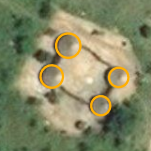} \\
  (a) & (b) & (c) \\  
  \includegraphics[width=0.25\columnwidth]{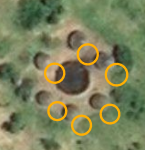} &
  \includegraphics[width=0.25\columnwidth]{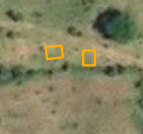} &
  \includegraphics[width=0.25\columnwidth]{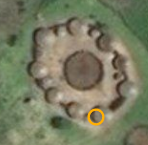} \\ 
  (d) & (e) & (f)
\end{tabular} 
\end{center}
\caption{
Examples of rural building annotations (orange circles), performed by volunteers in OpenStreetMap, superimposed over Bing imagery. a-c) correct annotation, d) annotations with misalignment errors, e) annotations that do not correspond to any building in the aerial images, f) incomplete annotations.
	\label{fig:osm_annotations}}
\end{figure}

In general, geographical objects in OSM are defined by their geometries (e.g., polygon) and  associated tags (e.g., building). In this work, we refer to OSM geographical objects as OSM \emph{annotations}. Figure~\ref{fig:osm_annotations} shows rural building annotations in OSM that were manually digitized by volunteers. Although some annotations are of good quality, several others have issues of misalignment or incompleteness with respect to aerial imagery.
Misalignment problems are present because of image co-registration issues. The incompleteness is due to new buildings that become visible in the updated imagery and also due to incomplete annotations that might have been submitted by volunteers. Additionally to these issues, modifications of the building can also result in geometry changes.

Deep learning methods are nowadays considered state-of-the-art techniques for several computer vision tasks, such as object detection and segmentation~\citep{He_2017_mask, Liu_2018_path}. These techniques have been also used successfully for building segmentation in aerial images~\citep{Maggiori_2017cnn, Saito_2016, Volpi_2017, Cheng_2019_darnet} and have led to accurate results in urban areas~\citep{Hamaguchi_2018, Wang_2017_torontocity}. These methods could be useful for automatic building extraction in large settlements\footnote{https://github.com/Microsoft/USBuildingFootprints}. However, the main drawback of deep learning methods is the requirement of a large amount of labeled data for training, usually in the range of several thousands of labeled samples. OSM data have been used for several applications including real-time routing~\citep{Luxen_2011}, autonomous driving~\citep{Fleischmann_2017navigation}, 3D building modeling~\citep{Wang_2017} and landuse classification~\citep{Srivastava_2018_finegrained, Srivastava_2019_multimodal}. 
~\citet{Audebert_2017} and~\citet{Kaiser_2017} used aerial images and OSM footprint annotations to train deep learning-based methods to perform semantic segmentation of buildings and road networks. 
More recently,~\cite{Chen_2018} proposed a deep learning-based building detection method that used data from three open crowdsourced geographic systems: OSM, MapSwipe~\footnote{https://mapswipe.org/}, and OsmAnd~\footnote{https://osmand.net/}. However, the two last resources just provided annotations at the tile level, useful for building patch detection but not for building segmentation. Active learning methods were proposed in~\cite{Chen_2018} and~\cite{Chen_2017} to select better image patches for annotation and then train an effective classifier from those patches. However, these works proposed solutions for patch-based building detection which cannot be directly used to automatically update maps of open mapping services, like OSM and Wikimapia~\footnote{http://wikimapia.org}, because those maps store vectorial footprints that delineate individual buildings.

Although OSM data has been used to train machine learning methods for some applications, the OSM annotations are very often of not good quality~\citep{Mnih_2012, Jilani_2014, Ali_2014, VargasMunoz_2019}. For instance,~\citet{Mnih_2012} found noisy labels (e.g., incomplete labels) of road networks and buildings that were used to create semantic segmentation methods.
~\cite{Jilani_2014} observed that there exist errors in the tags associated with different types of road networks in OSM data, and proposed a method for automatic road tagging. In other works,~\cite{Ali_2014, Ali_2017} found several green areas of different classes (e.g., meadows, grass, parks, and gardens) that have incorrect labels in OSM. In the specific case of building annotations,~\cite{VargasMunoz_2019} identified three main issues in building annotations of OSM: i) location inaccuracies (see an example in Figure~\ref{fig:osm_annotations}d), ii) annotations not related to a building (see Figure~\ref{fig:osm_annotations}e) and iii) missing annotations (see Figure~\ref{fig:osm_annotations}f). 

In addition to the problem of low data quality in OSM, the amount of available labeled data is small in rural areas as compared to urban regions. This happens because the number of volunteers that update OSM data drops outside cities~\citep{Neis_2014}: usually these areas get attention of the volunteers only when an epidemy or a natural disaster happens there. This makes more difficult to train accurate models for the segmentation of buildings in rural areas. \cite{VargasMunoz_2019} proposed a methodology to automatically correct rural building annotations with the aforementioned problems in OSM. However, the accuracies attained by this method are not sufficient for production-ready automatic updates of OSM. 

An additional problem for the methods that try to automate the process of updating automatically OSM building data is that the output of such methods should be vectorial building footprints. Recently,~\cite{Marcos2018cvpr} proposed a method that outputs vectorial building delineations using a method based on Active Contour Models (ACM), which uses a Convolutional Neural Network (CNN) to learn the parameters of the ACM model. A different approach is proposed in~\cite{Tasar_2018} via a mesh-based approximation method that converts binary building classification maps into polygonized buildings. \cite{VargasMunoz_2019} proposed a CNN that detects rural buildings of predefined shapes, which can be easily exported to vectorial format. All these methods strongly depend on the accuracy of the previously computed building map.

In this work, we propose a methodology for interactive correction of rural buildings in OSM. Our proposed method starts by correcting misalignment errors of the annotations using a Markov Random Field (MRF) based method proposed in~\cite{VargasMunoz_2019}. We selected that approach, because it  outperforms other methods for the particular task of building footprint alignment. From that moment on, the model interacts with a human operator for digitizing objects in OSM. The model selects the image locations that maximize the chances of editing corrections.  

In this way, the user just analyzes a small set of selected regions of a large geographical area. Our aim is to reduce the effort of the user for mapping buildings in a certain geographical area, by intelligently selecting just the regions that require user corrections/annotations. The selection of these regions is determined by analyzing a building probability map obtained by a CNN method on aerial images and the current OSM annotations. Given that large areas need to be analyzed to select the most interesting regions for annotation, we propose an efficient CNN model for building segmentation that performs fast inference. To do so, we enrich our CNN with a branch for early stopping when there are no buildings in the analyzed image patch. Whenever a considerable amount of new labeled data is available, the CNN model is retrained to improve the accuracy of the building predictor, and so the accuracy of the selection of tiles that need annotations/corrections. Experiments performed by simulation and real human annotations in datasets of different countries (four different datasets) show 
that the proposed approach effectively reduces the number of tiles that need to be screened and therefore the human workload. In most of our experiments, we found that our method allows to correct 98\% of the wrong annotations by analyzing less than a quarter of the data.

In Section~\ref{sec:methodology}, we present the proposed methodology for interactive annotation/correction of OSM rural building annotations. Section~\ref{sec:data_and_experimental_setup} presents the data set and the setup used in our experiments, and Section~\ref{sec:results} shows the experimental results. Finally, Section~\ref{sec:conclusions} presents the conclusions of the paper.

\section{Methodology}
\label{sec:methodology}

\begin{figure}[p]
\centering
\includegraphics[width=0.9\columnwidth]{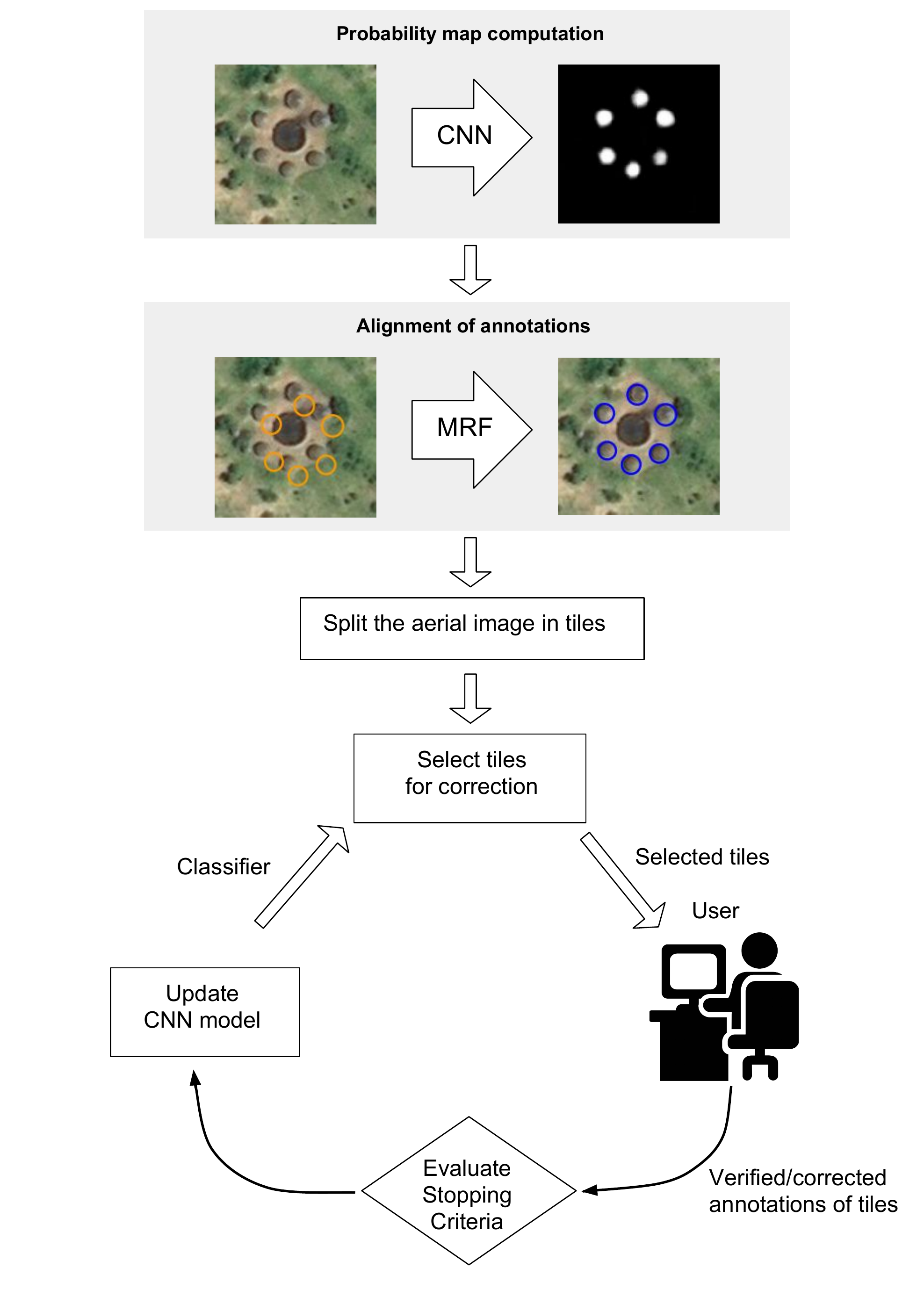}
\caption{The proposed methodology for interactive correct OSM rural building annotations. The orange circles represent misaligned OSM building annotations while the blue circles represent the annotations after the misalignment correction.\label{fig:methodology}}
\end{figure}

Our proposed methodology is illustrated in Figure~\ref{fig:methodology}. First, we train a CNN building segmentation model to obtain a building probability map, which is used to perform the alignment of the original OSM polygons. Then, the image being analyzed is split into tiles, in order to select a small number of regions that need manual editing in OSM. This is done based on the probability map and the currently aligned polygons. After the annotation of the selected samples, a stopping criterion is verified to finish the process of annotation. If this criterion is not met, the new building annotations are used to improve the building classification CNN, which in turn improves the process of selection of tiles in the next iteration. The next sections explain in detail the steps of the proposed methodology.

\subsection{Computation of the building probability map}
\label{subsec:compute_probmap}

\begin{figure}[!t]
\begin{center}
\begin{tabular}{c} 
  \includegraphics[width=1.0\columnwidth]{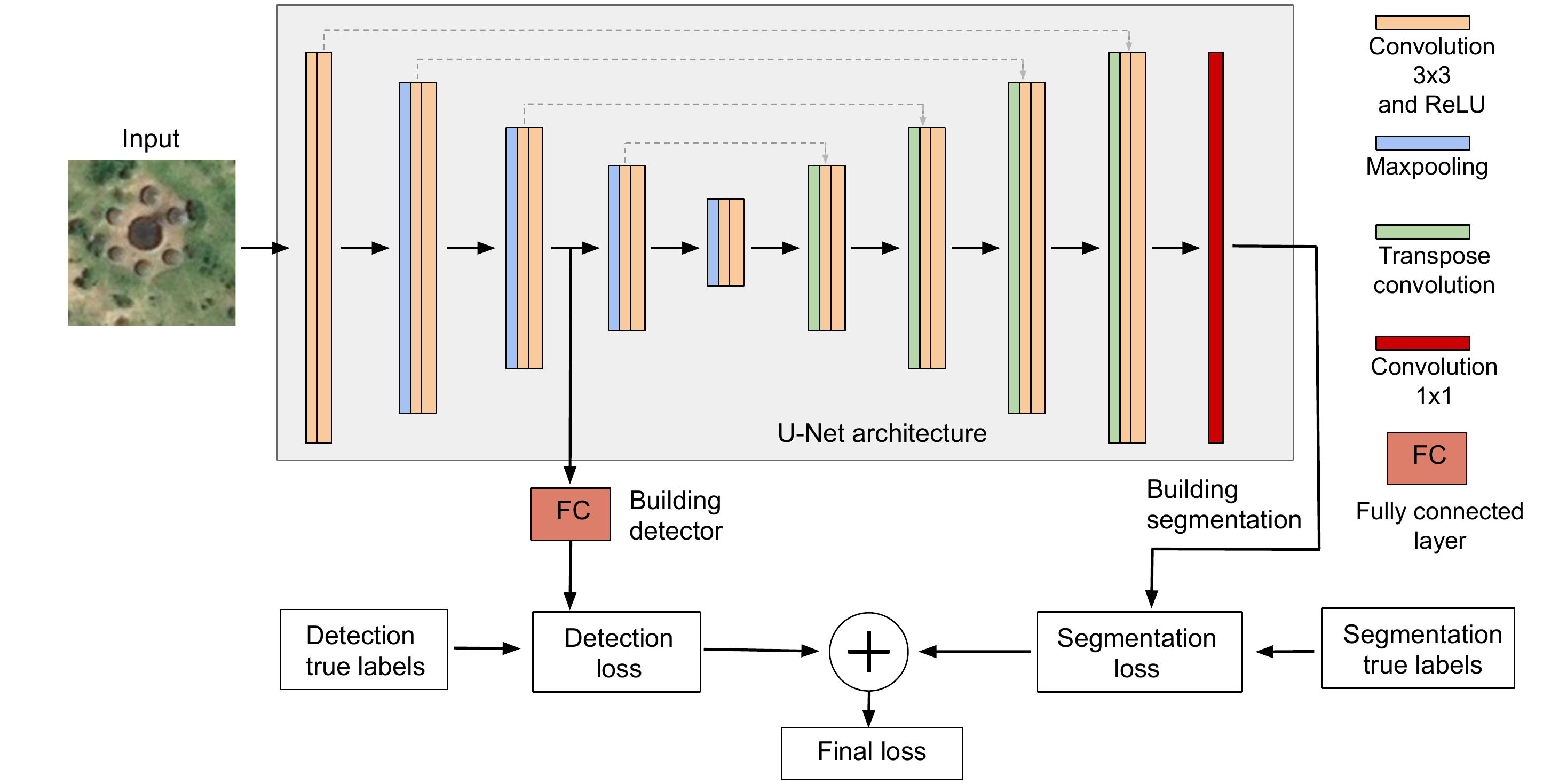} \\
  (a) Loss computation \\  
  \\
  \hline
  \\
  \includegraphics[width=1.0\columnwidth]{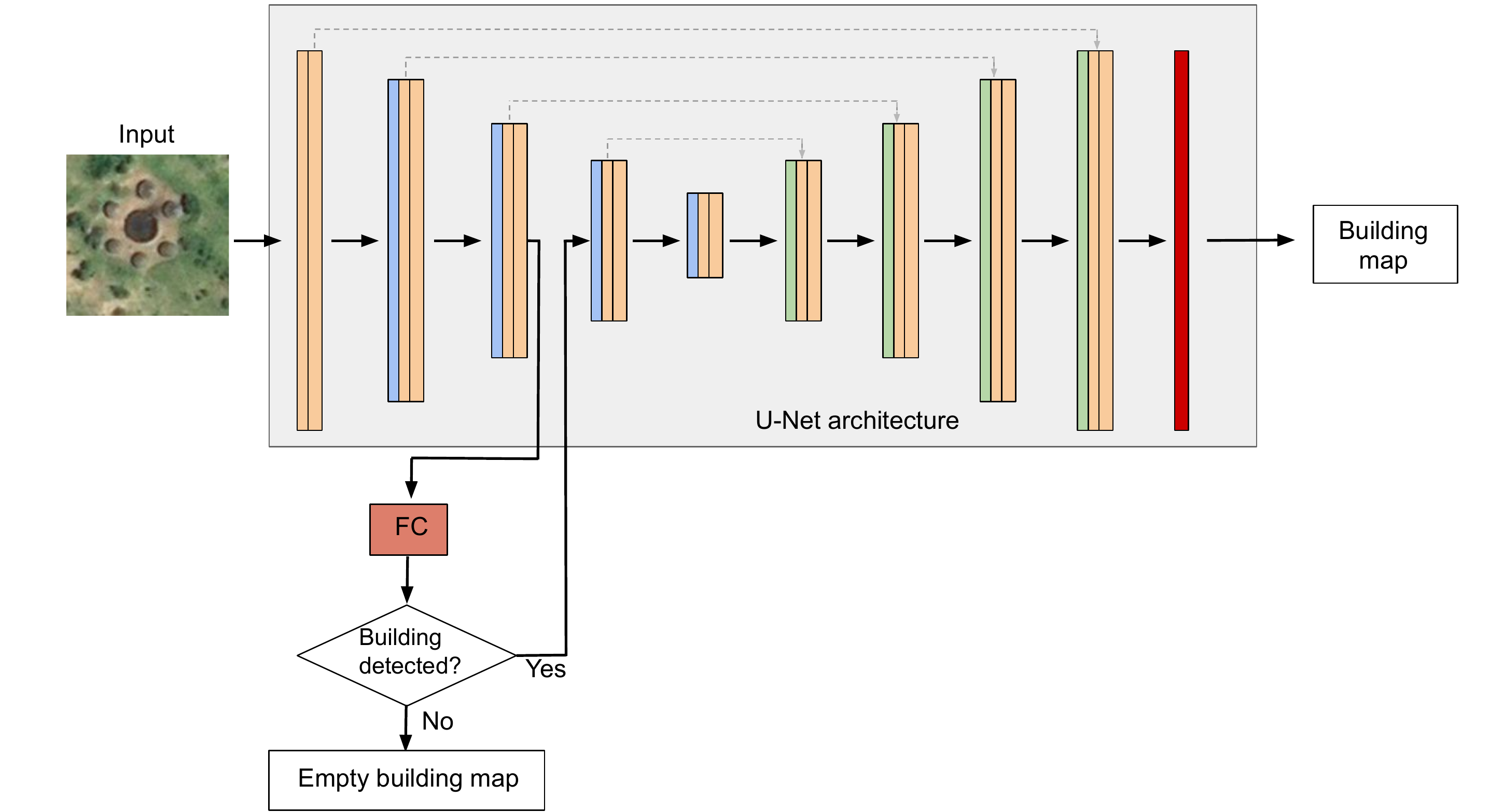}\\
  (b) Proposed strategy for fast inference
\end{tabular} 
\end{center}
\caption{Proposed method for building segmentation: a) Loss computation based on a detection and segmentation loss b) Fast inference by using a early exiting brach. 
	\label{fig:proposed_fast_building_segmentation}}
\end{figure}

Given that our model is applied to large areas where rural buildings are sparsely located, we propose an efficient CNN-based method for building segmentation. Our proposed model is based on the U-Net architecture proposed in~\cite{Ronneberger_2015}. U-Net has shown good performance in several applications and recently an ensemble of U-Net models has obtained the best performance in a building segmentation competition, the DeepGlobe challenge~\cite{Dem18}. The U-Net model (illustrated in the gray boxes of Figure~\ref{fig:proposed_fast_building_segmentation}) consists of several operations that extract image features and capture contextual information, alongside a symmetric set of operations that upsample the feature maps, therefore enabling precise pixel-level semantic labeling. We extend the U-Net model by adding an extra branch after the third set of convolution groups (see Figure~\ref{fig:proposed_fast_building_segmentation}a). This branch of the model considers that the image features extracted until that point are sufficient to identify whether the image contains buildings. A fully connected layer, denoted as $FC$, is applied over the aforementioned features to perform binary classification (i.e., there are buildings in the patch or not). In order to train the model with the additional branch, we define a new loss function illustrated in Figure~\ref{fig:proposed_fast_building_segmentation}b. Our loss function is defined as the sum of a detection and segmentation loss, both computed using binary cross entropy:

\begin{eqnarray}
L_{seg}(\hat{y}^s, y^s) &=& - \frac{1}{N} \sum_{i}^{N} y^s_i  \log(\sigma(\hat{y}^s_i)) + (1-y^s_i) \log(1 - \sigma(\hat{y}^s_i))\\
L_{det}(\hat{y}^d, y^d) &=& - \frac{1}{M} \sum_{j}^{M} y^d_j  \log(\sigma(\hat{y}^d_j)) + (1-y^d_j) \log(1 - \sigma(\hat{y}^d_j))\\
L(\hat{y}^s, \hat{y}^d, y^s, y^d) &=& L_{seg}(\hat{y}^s, y^s) + L_{det}(\hat{y}^d, y^d),
\label{eq:final_loss_loss}
\end{eqnarray}

where $\sigma$ is the sigmoid function, $\hat{y}^s$ is the output of the last convolutional layer of the U-Net model and $y^s$ is the segmentation ground truth map, $\hat{y}^d$ is the detection prediction obtained from $FC$ and $y^d$ is the detection ground truth, $N$ is the number of samples in the processed batch of pixels (segmentation output) and $M$ is the number of samples in the processed batch of images.

During inference, the proposed model computes sequentially the first three groups of convolutions  followed by a fully connected layer and a sigmoid function that outputs the building presence probability value. If this value is higher than a certain threshold $\theta$, then we consider that there exist buildings in the analyzed patch and the rest of the U-Net model is executed to obtain a building segmentation map. Otherwise, we consider that there are no buildings in the image patch and the model outputs 
an empty probability map. In this way, the inference time is reduced.

\subsection{Correction of misalignment errors}
\label{subsec:correction_of_misalignement_errors}

In order to correct alignment errors in OSM, we use the approach proposed in~\cite{VargasMunoz_2019} that showed good performances for the alignment of rural building annotations. In that work, the OSM polygons of rural buildings are grouped based on spatial proximity. Then, groups of buildings are aligned with a single shift vector. Aligning groups of polygons makes the method more robust to building probability maps of bad quality. The correlation of the aligned annotations and building probability maps is used to measure the performance of a given alignment vector. Given that nearby groups of buildings have similar registration errors,~\cite{VargasMunoz_2019} used MRF to find the alignment vectors $\mathbf{d} = \{d_0, d_1, \ldots, d_n\}$ that can correct the current annotation locations $\mathbf{x}$ using a building probability map $\mathbf{y}$. The optimum alignment vectors minimize the following energy function:

\begin{align}
\mathbf{\hat{d}} =\arg\min\limits_\mathbf{d\in\mathbf{v}^\mathcal{N}} \sum\limits_i -\log C(A(x_i,d_i), y_i) + \beta \sum_{j \in \mathbf{h}_i} \frac{1}{\mathcal{Z}} ||d_i - d_j||_2,\label{eq:energy_total_interactiveosm}
\end{align}
where the unary term, that measures the matching between the annotations and probability map, is computed by using the normalized dot product $C(A(x_i,d_i), y_i)$ between the aligned annotations $A(x_i,d_i)$ ($A$ is a function that shifts the position of the annotation $x_i$ using the alignment vector $d_i$) and the probability map $y_i$, while the pairwise term is obtained by computing the vector norm of the difference of the alignment vector $d_i$ and the neighboring alignment vectors $d_j$ ($j \in \mathbf{h}_i$). In equation~\ref{eq:energy_total_interactiveosm} $\mathbf{v} = \{ v_1, v_2, \ldots , v_p \}$ represents the set of all the alignment vectors, $\mathcal{Z}$ is a normalization term computed as the maximum possible distance between two alignment vectors in $\mathbf{v}$ and $\beta$ is a regularization parameter. We use the Iterative Conditional Modes (ICM) algorithm to find the solution of equation~\ref{eq:energy_total_interactiveosm}.

\subsection{Selection of tiles for correction/verification}
\label{subsec:selection_of_tiles}

Once the building probability map has been generated and the existing OSM annotations have been aligned to it, volunteers can start the editing process in OSM. The aim of this section is to define a criterion to focus the volunteers' efforts where it really matters (regions with wrong or missing annotations). In this respect, it is crucial to obtain a measure of how incorrect an annotation is, based on the current OSM annotations and a building probability map. Such measure is used to create an ordered list of tiles $L$, where those whose annotations have more chances to be incorrect are ranked higher. Then, the top $t$ tiles from $L$ are selected for correction/verification by a user. We convert the OSM vectorial polygons to images with value one in the positions that are inside a building annotation, otherwise, zero everywhere else. This is done in order to be able to compare the OSM annotations with the building probability map. In the next sections, we describe the measures proposed.

\vspace{.2cm}
\noindent{\textbf{Mutual information}}

Mutual information (MI) has been used as an effective similarity metric to compare images~\cite{Pluim_2003}. The MI of two variables measures the amount of information that one variable carries about the other. The MI of two images $A$ and $B$ can be defined as follows:

\begin{equation}
MI(A;B) = \sum_{b \in B} \sum_{a \in A}
p(a,b) \log{ \left(\frac{p(a,b)}{p(a)\,p(b)} \right), }\label{eq:mutual_info}
\end{equation}
where $p(a,b)$ is the joint probability distribution of the pixels that correspond to $A$ and $B$ and $p(a)$ and $p(b)$ are the marginal probability distributions of $A$ and $B$, respectively.
For the MI metric, the lower the value the higher the priority to be selected for verification/correction.

\vspace{.2cm}
\noindent{\textbf{Normalized dot product}}

\cite{VargasMunoz_2019} used the normalized dot product (NDP) to measure the degree of matching between the building annotations and the building probability map. This measure is defined as follows:

\begin{equation}
NDP(A;B) = \frac{A \cdot B}{size(A)},\label{eq:norm_dot_prod}
\end{equation}
where $A$ and $B$ are images of the same size, ($\cdot$) represents the dot product operation and $size(A)$ is the number of pixels in image $A$.
For the NDP metric, the lower the value the higher the priority to be selected for verification/correction.

\vspace{.2cm}
\noindent{\textbf{Sum of absolute differences}}

The sum of absolute differences (SAD) of the pixel values of two images is an efficient and effective way to measure the degree of matching of two images, as shown in~\cite{Alsaade_2012}. 
For the SAD metric, the higher the value the higher the priority to be selected for verification/correction.

\subsection{User annotation and evaluation of the stopping criteria}
\label{subsec:user_annotation}

The user verifies/corrects the annotations in the selected tiles. In order to correct annotations, the user applies three types of manual operations that are available in a Graphical User Interface (GUI), for example, the web-based iD editor~\footnote{http://ideditor.com/} of OSM:  align (drag and drop annotations into the right position), remove and add (digitize manually new polygons). The user performs these operations by visually inspecting the original OSM annotations superimposed over the aerial imagery.

After the user verification/correction of the selected tiles, a stopping criterion is evaluated. We propose as criterion the percentage of tiles that required correction since the last $k$ analyzed tiles, denoted as $p_k$. This measure is updated after every tile is analyzed by the user. Note that the tiles are selected from a list $L$, sorted in ascending order in terms of expected correctness of the annotations available, estimated with measures presented in Section~\ref{subsec:selection_of_tiles}. Thus, the more tiles are analyzed by the user, the lower the value of $p_k$. This happens because, eventually, the majority of the tiles with incorrect annotations will be already analyzed and the majority of the remaining tiles will not need any correction. Once $p_k$ is lower than a given threshold $r_k$ the process of user verification/correction ends.

After a considerable amount of tiles (e.g., 20) are verified/annotated by the user, the proposed CNN building segmentation model is fine-tuned with the new annotated data. This updated model is then used to improve the building probability maps, which in turn improves the performance of the annotation correctness measures described in Section~\ref{subsec:selection_of_tiles}.

\section{Data and experimental setup}
\label{sec:data_and_experimental_setup}

\subsection{Datasets} 

For the validation of our proposed methodology, we used data obtained from the countries of the United Republic of Tanzania and the Republic of Zimbabwe. The CNN building segmentation model was trained with $3134$ OSM rural buildings annotations from several regions of Tanzania. These OSM annotations for these regions were manually verified/corrected on a set of Bing aerial images, that cover $23.75$ km$^2$, with $30$cm spatial resolution, acquired over the Geita, Singida, Mara, Mtwara, and Manyara regions of Tanzania.

To evaluate our proposed methodology with simulations of user annotation corrections we collected two test datasets, spatially disjoint with respect to the training set. The first dataset is collected in the country of Tanzania and is composed of $1267$ OSM building footprints (in an area of 25.5 km$^2$) that contain annotation errors. These annotations were located close to the region of Mugumu in Tanzania, this data is called the Tanzania dataset in Section~\ref{sec:results}. The second dataset contains $1392$ OSM annotations (in an area of 27.9 km$^2$), that also contains errors, is collected in the region of Midlands in Zimbabwe, and is called Zimbabwe dataset in Section~\ref{sec:results}. In order to perform simulations of user annotations, we manually corrected the annotations in these two datasets.

In order to perform experiments with real user interactions, we used OSM annotations from two regions. The first one located close to Mugumu region in Tanzania, denominated as Mugumu dataset (646 building footprints collected from an area of 9.4 km$^2$). The second evaluation area is located in the Gweru region, in Zimbabwe, denominated as Gweru dataset (1273 building footprints collected from an area of 28.4 km$^2$).

\subsection{Model setup}
\label{subsec:model_setup}

The CNN model was trained for $20$ epochs with an initial learning rate of $0.001$ decreased by a factor of $0.1$ after every $10$ epochs. During the interactive annotation process, we finetuned the CNN model for $10$ additional epochs and a learning rate of $0.0001$ with the new annotated data. 
For our proposed CNN model, we used $\theta = 0.1$ to predict if an image patch contains or not buildings. As in~\cite{VargasMunoz_2019}, we used $\beta = 2.0$ for the alingment correction process.
We split the images corresponding to the selected geographical regions in tiles of size $256 \times 256$ pixels. This tile size was chosen because we observed that it is large enough to cover groups of rural buildings with little background information. In order to evaluate the proposed stopping criteria, we fixed the parameter value $k = 100$ and evaluated the performance of different values $r_k \in \{0.02, 0.05, 0.10\}$. 
For the experiments performed to choose the tile selection strategy, we took into consideration a common tricky case that affects the metrics MI and NDP. In the case where there are no annotations in a tile but there are a small amount of false positive predictions, the aforementioned metrics will output zero. Then, the method will assign a high priority to the tiles to be verified/corrected by a user, which is not correct since there is nothing to edit in that tile. Thus, for MI and NDP, we verified if the tile does not contain OSM annotations, and if the CNN prediction has a building with size shorter than a very small building in our dataset (20 pixels of size). If that is the case we output the value one which will assign a low selection priority to the tile.

\subsection{Setup for experiments with real user annotation corrections}
\label{subsec:setup_real_users}

\begin{figure}[!t]
\begin{center}
\begin{tabular}{cc} 
  \includegraphics[width=0.4\columnwidth]{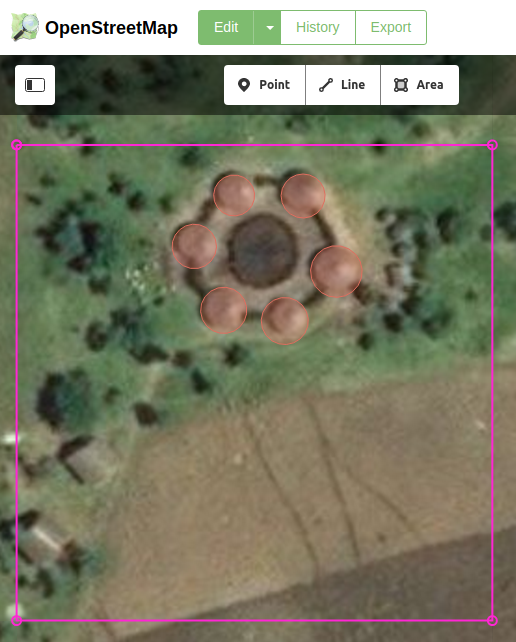} &
  \includegraphics[width=0.4\columnwidth]{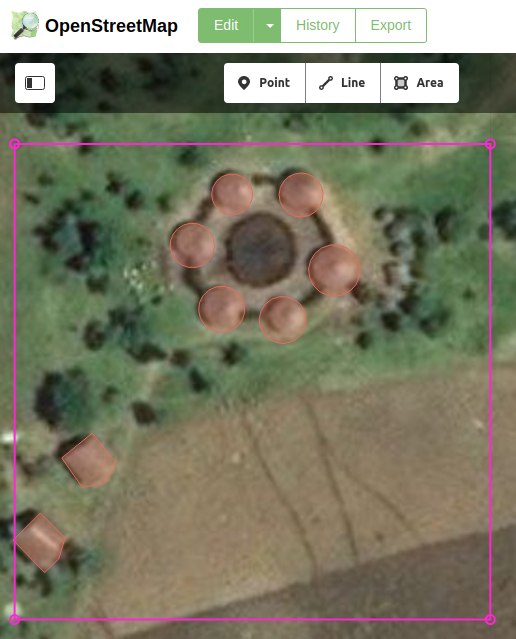} \\ 
  (a)  & (b) \\
\end{tabular} 
\end{center}
\caption{Gaphical User Interface of the iD editor for editing OpenStreetMap annotations: a) Tile (deliminted by the bounding box in magenta) loaded for verification/correction of rural building annotations, b) OSM map after the annotation of two new buildings in the bottom left part of the tile.\label{fig:id_editor}}
\end{figure}

So far, the proposed methodology has considered simulated annotation corrections to obtain perfect annotations. 
In order to evaluate our proposed methodology in a realistic scenario, we also performed a set of experiments with real volunteers performing annotations. To run those tests we deloped a web application that allows several users to correct annotations in OpenStreetMap, by using the iD editor~\footnote{http://ideditor.com/} API. Given a geographical area where we want to correct the rural building annotations, our method selects the tiles that need to be verified/corrected by the user with higher priority. When a user asks for a tile for analysis, the web application loads a tile highlighted with a bounding box in magenta in the iD editor. Figure~\ref{fig:id_editor}a depicts the bounding box of the selected tile superimposed over aerial imagery and the current rural building annotation at that location. The user is asked to correct/verify existing annotations, and also to add new building annotations if needed. For instance Figure~\ref{fig:id_editor}b shows the annotations after user corrections, where two missing buildings were added on the bottom left of the bounding box. For these experiments, 9 volunteers used the web application to interactively correct OSM rural building annotations, in March 2019. The volunteers were students of the Computer Science department, at the University of Campinas, Brazil. They were unfamiliar with OSM and the digital annotation tasks. After an introduction session to the editing environment, they were asked to perform around 7 hours of annotations in two working sessions (labeling the Mugumu and Gweru dataset).

It is difficult to ensure that the new annotations are of good quality. This depends a lot on the volunteers' skills. Thus, it was very important to perform an introductory tutorial of the building editing tool (i.e., the iD editor of OSM). Such quality checks can be implemented in real deployments of the system (e.g. using golden questions or evaluating users' behaviour (e.g.~\citet{Gom11}), but this goes beyond the scope of this paper and will be considered in future research.

\section{Results}
\label{sec:results}

In this section, we present the results obtained by the proposed methodology. Section~\ref{subsec:evaluation_of_proposed_cnn} compare the proposed CNN model with the standard U-Net model.
Using the computed probability maps obtained by our proposed CNN method we correct alignment errors, as explained in Section~\ref{subsec:correction_of_misalignement_errors}. Section~\ref{subsec:alignment_results} evaluates the alignment results. Then, after correcting existing alignment errors 
in Section~\ref{subsec:results_selection_strategy} we compare several approaches to select the tiles that require correction. Section~\ref{subsec:results_stopping_criteria} evaluates the proposed stopping criteria by analyzing the percentage of wrong annotations corrected and the percentage of analyzed tiles when the stopping criteria is met. In Section~\ref{subsec:real_user_experiments}, we report the results of the experiments including real annotation corrections. 

\subsection{Evaluation of the proposed CNN method to compute a building probability map}
\label{subsec:evaluation_of_proposed_cnn}

We compared the efficiency and efficacy of the proposed CNN method with the standard U-Net model, in the Tanzania dataset. The computed probability maps were thresholded to obtain the pixel-based binary classification map (i.e., greater than 0.5 building pixel otherwise background pixel). For the comparison, we used precision, recall, and F1-score at the object-level, as in~\cite{VargasMunoz_2019}. Both models were trained with the configuration explained in Section~\ref{subsec:model_setup}.  Table~\ref{tab:proposed_vs_unet} shows that the proposed method attains better performance than U-Net, in the three metrics. It is also more than 70\% faster than the U-Net model, reducing the processing time from 157 to 90 secs in the evaluated dataset.
Our proposed method performs first detection (classifying if an image patch contains buildings or not) and performs semantic segmentation (pixel-based classification) just if needed.
It might happen that some image patches classified as not containing buildings actually contain buildings and semantic segmentation should be applied to them. Thus, we measure the detection recall to verify that our model is not having many errors of this type. Table~\ref{tab:proposed_vs_unet} shows the detection recall of our proposed CNN method, which attains a very high value.

\begin{table}[!t]
\centering
\caption{Performance of the proposed CNN method for building detection and segmentation as compared to the standard U-Net model. 
}
\label{tab:proposed_vs_unet}
\begin{tabular}{|l|r|r|r|r|r|}
\hline
Methods &\multicolumn{5}{c|}{Tanzania dataset}  \\
\cline{2-6}
 &\multicolumn{1}{c|}{Precision}&\multicolumn{1}{c|}{Recall}&\multicolumn{1}{c|}{F1-score}&\multicolumn{1}{c|}{Time(sec)}&\multicolumn{1}{c|}{Detection recall} \\

\hline
Proposed & 0.47 & 0.84 & 0.60 & 90.0 & 0.99 \\ 
U-Net  & 0.43 & 0.72 & 0.54 & 157.0 & - \\ 
\hline
\end{tabular}
\end{table}

\subsection{Evaluation of the alignment method}
\label{subsec:alignment_results}

\begin{figure}[!t]
\begin{center}
\begin{tabular}{ccc} 

\multicolumn{3}{c}{Example 1}\\
  \includegraphics[width=0.20\columnwidth]{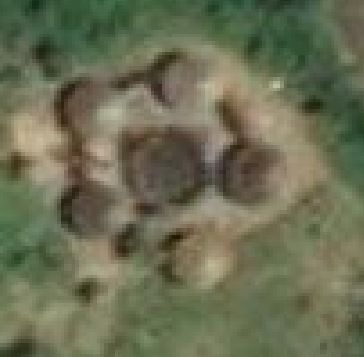} & 
  \includegraphics[width=0.20\columnwidth]{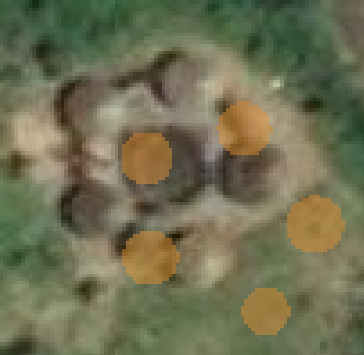} &  
  \includegraphics[width=0.20\columnwidth]{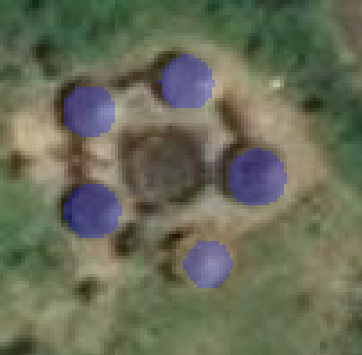} \\
  (a) Image & (b) Original annotations & (c) Aligned annotations \\
   \hline
   
  \multicolumn{3}{c}{Example 2}\\
  \includegraphics[width=0.20\columnwidth]{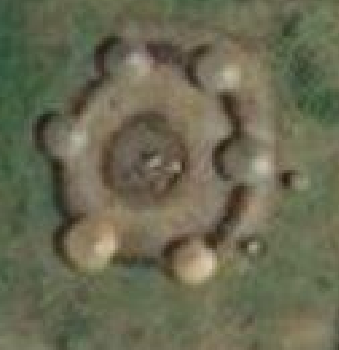} & 
  \includegraphics[width=0.20\columnwidth]{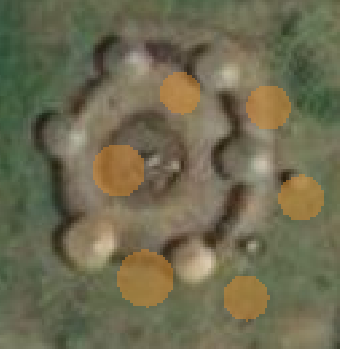} &  
  \includegraphics[width=0.20\columnwidth]{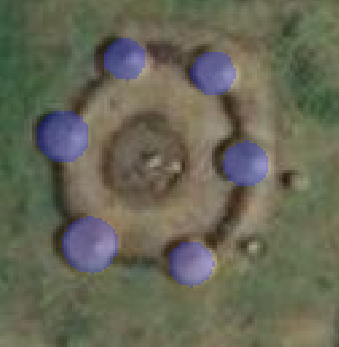} \\
  (d) Image & (e) Original annotations & (f) Aligned annotations \\
      \hline
 
  \multicolumn{3}{c}{Example 3}\\
  \includegraphics[width=0.20\columnwidth]{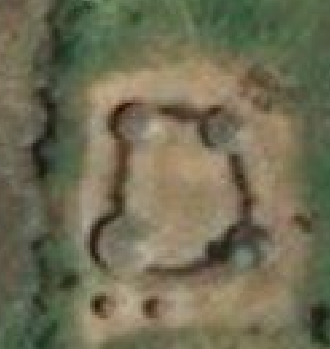} & 
  \includegraphics[width=0.20\columnwidth]{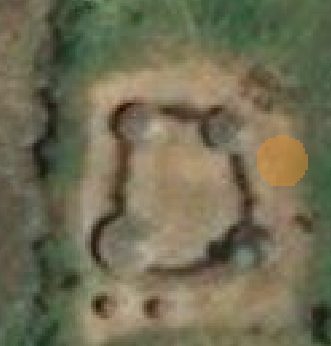} &  
  \includegraphics[width=0.20\columnwidth]{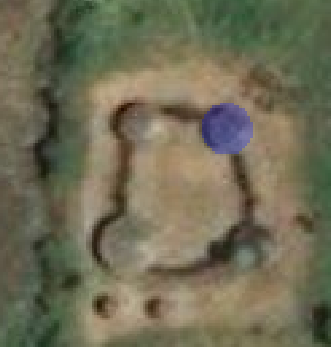} \\
  (g) Image & (h) Original annotations & (i) Aligned annotations \\
\hline

  \multicolumn{3}{c}{Example 4}\\
  \includegraphics[width=0.20\columnwidth]{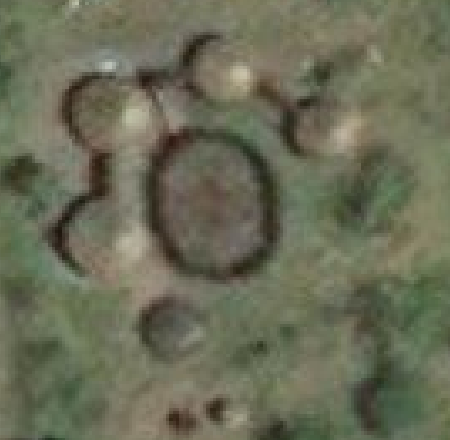} & 
  \includegraphics[width=0.20\columnwidth]{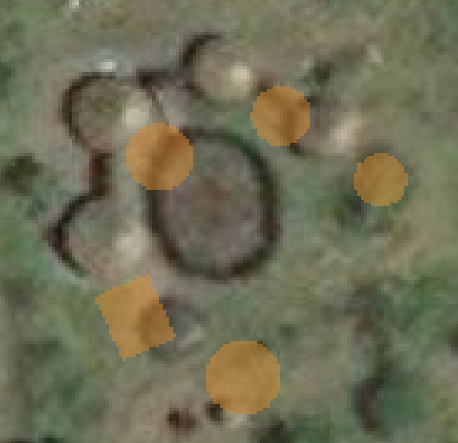} &  
  \includegraphics[width=0.20\columnwidth]{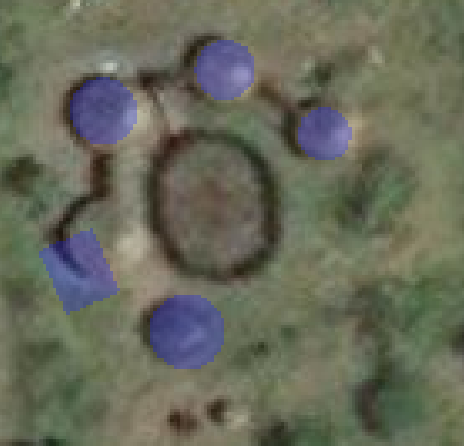} \\
  (j) Image & (k) Original annotations & (l) Aligned annotations \\
\end{tabular} 
\end{center}
\caption{Examples of alignment results in the Tanzania dataset. The original misaligned annotations are presented in orange and the aligned annotations in blue.
\label{fig:visual_alignment_examples}}
\end{figure}

We applied the alignment method, presented in Section~\ref{subsec:correction_of_misalignement_errors}, to the original OSM annotations of the Tanzania dataset. 
In order to measure how well this method performs we computed the object-level accuracy of overlapping annotations with the ground-truth. This is computed as the number of annotations that have a strong overlap with a ground-truth annotation, Intersection over Union (IoU) greater than 0.5 as in~\cite{VargasMunoz_2019}, divided by the number of annotations that have at least a very small overlap with a ground-truth annotation (IoU greater than 0.05). The object-level overlapping accuracy of the aligned annotations is 93.9\%, which is much better than 12.5\% obtained by the original OSM annotations (without alignment). Figure~\ref{fig:visual_alignment_examples} presents visual examples of OSM annotations before and after alignment. The first two rows in Figure~\ref{fig:visual_alignment_examples} show examples that by aligning the annotations we obtain a correct set of annotations in those locations. The third row shows an example where there is just one OSM annotation when there are many buildings in the aerial imagery. Thus, after applying alignment there are still buildings that need to be digitized by annotators. The fourth row presents an example where all the buildings seem to have a circular shape but the leftmost annotation have a rectangular shape. Thus, after performing alignment the annotator will still need to remove the polygon with wrong shape and digitize a new building annotation.

\subsection{Tile selection strategy}
\label{subsec:results_selection_strategy}

\begin{figure}[t]
\centering
\includegraphics[width=0.8\columnwidth]{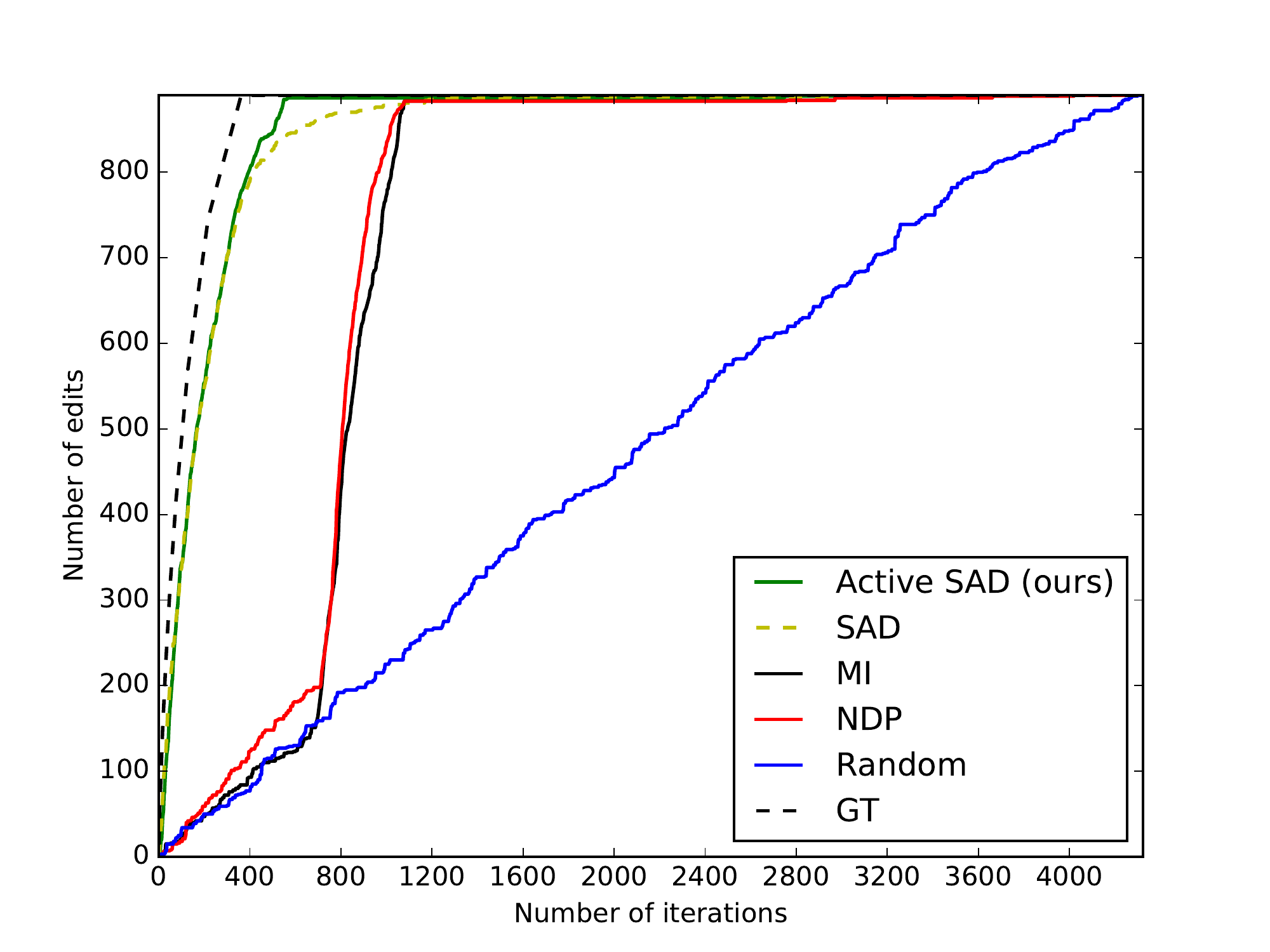}
\caption{Results on the Tanzania dataset of our proposed methodology for interactive annotation of rural buildings in OSM using several measures for error annotation detection. The vertical axis of the plot shows the number of annotation errors detected by the evaluated strategy and the horizontal axis represents the number of tiles analyzed by the user.\label{fig:plot_tanzania_dataset}}
\end{figure}

\begin{figure}[t]
\centering
\includegraphics[width=0.8\columnwidth]{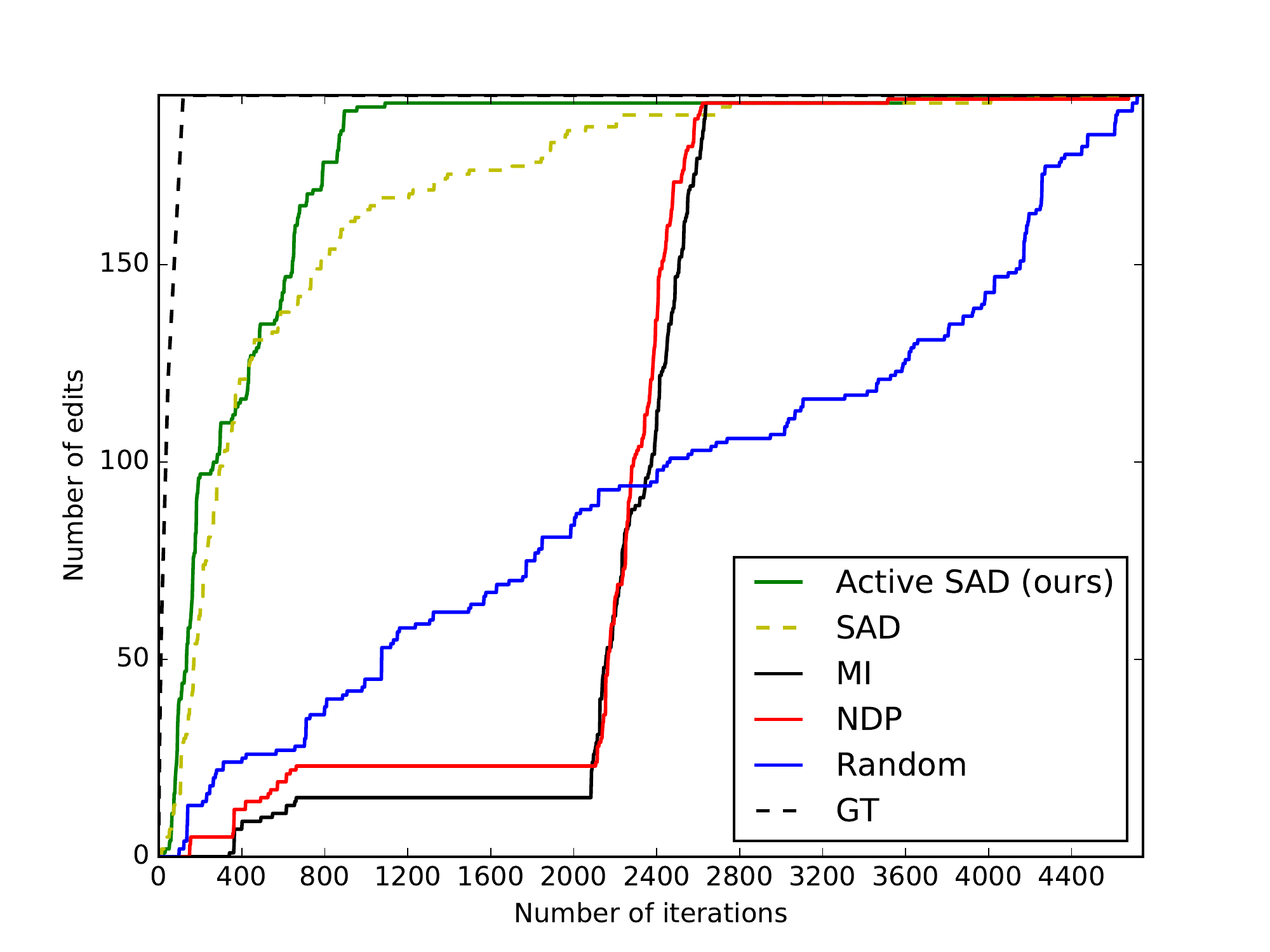}
\caption{Results on the Zimbabwe dataset of our proposed methodology for interactive annotation of rural buildings in OSM using several measures for error annotation detection. The vertical axis of the plot shows the number of annotation errors detected by the evaluated strategy and the horizontal axis represents the number of tiles analyzed by the user.\label{fig:plot_zimbabwe_dataset}}
\end{figure}

After correcting the alignment of the original OSM annotations, we evaluated the different selection strategies presented in Section~\ref{subsec:selection_of_tiles}, to find which approach can better rank the tiles that require corrections, and therefore minimize the effort of the user to verify/correct rural building annotations in OSM. 
We included as a baseline a method that randomly selects tiles to be annotated by the user and also show an upper bound that uses the ground-truth (GT) data to select the tiles. Figure~\ref{fig:plot_tanzania_dataset} shows the results of our proposed methodology in the Tanzania dataset using different strategies to select tiles for annotation: MI, NDP and SAD. We also show a strategy, called Active SAD, that uses the SAD to select tiles and then updates the CNN segmentation model with the new verified/annotated data (i.e., the method using the complete interactive pipeline in Figure~\ref{fig:methodology}). We can observe that SAD performs considerably better than MI and NDP and Active SAD performs better than SAD in the last iterations when a great part of the tiles with errors have been discovered and corrected. Figure~\ref{fig:plot_zimbabwe_dataset} shows the results of the evaluated methods in the Zimbabwe dataset. The same trend can be observed in this dataset, SAD performs better than the other selection methods. However, in this dataset Active SAD is considerably better than SAD. This happens because the building probability maps of the Zimbabwe dataset are less accurate than the probability maps obtained for the Tanzania dataset. Therefore updating the building segmentation CNN model has more impact on the improvement of the selection strategy.
Retraining the model with the new corrected annotations adapts the model to the dataset and reduces domain adaptation problems~\cite{Matasci_2012}.

\subsection{Stopping criteria}
\label{subsec:results_stopping_criteria}

We evaluated the stopping criteria in the Tanzania and Zimbabwe datasets.
For the experiments, we used the method Active SAD, since it showed the best performance among the other analyzed tile selection approaches (see Section~\ref{subsec:results_selection_strategy}).
Table~\ref{tab:stopping_criteria} shows the percentage of wrong annotations corrected and the percentage of tiles analyzed when the stopping criterion is met.
Note that \% corrected is computed as the number of wrong annotations shown for user correction  divided by the total number of wrong annotations. 
We can observe that in the Tanzania dataset, when the annotation process stops, most of the wrong annotations were already corrected and around 15\% of the tiles were analyzed for all the three values of $r_k$. In the Zimbabwe dataset $r_k = 10\%$ stops too early the annotation process, and leads to the correction of just around half of the wrong annotations. Using $r_k = 2\%$, almost all the wrong annotations are detected but $20.9 \%$ of the tiles are analyzed. The difference in the results in the two datasets is mainly due to the difference in the accuracy of the respective building probability maps.

\begin{table}[!t]
\centering

\caption{Percentage of wrong annotations corrected (\% corrected) and percentage of tiles analyzed for three different values of $r_k$ (stopping criteria parameter) in the Tanzania and Zimbabwe datasets.}
\label{tab:stopping_criteria}
\begin{tabular}{|l|r|r|r|r|}
\hline
$r_k$ (\%) &\multicolumn{2}{c|}{Tanzania dataset} &\multicolumn{2}{c|}{Zimbabwe dataset} \\
\cline{2-5}
 &\multicolumn{1}{c|}{\% corrected}&\multicolumn{1}{c|}{\% tiles analyzed}  
 &\multicolumn{1}{c|}{\% corrected}&\multicolumn{1}{c|}{\% tiles analyzed} \\
\hline
10 & 99.7 & 14.8 & 52.8 & 6.0 \\ 
5  & 99.7 & 14.9 & 87.6 & 16.3 \\ 
2  & 99.7 & 15.0 & 98.4 & 20.9 \\ 
\hline
\end{tabular}

\end{table}

We also evaluated the behavior of the stopping criteria in datasets that have annotations of different quality levels. To do that we validated the performance of the proposed heuristic in several datasets obtained by simulations of removal, addition, and random small shifts of buildings annotations. For a given initial set of annotations, we randomly add and remove a certain number of annotations and then randomly shift the annotations in the horizontal and vertical axis in the range of 0 to 2 pixels. We performed these operations over the Tanzania and Zimbabwe datasets by adding and removing a percentage of the initial number of annotations.

\begin{figure}[!t]
\begin{center}
\begin{tabular}{cc} 
\hline
\multicolumn{2}{c}{Tanzania dataset}\\
  \includegraphics[width=0.5\columnwidth]{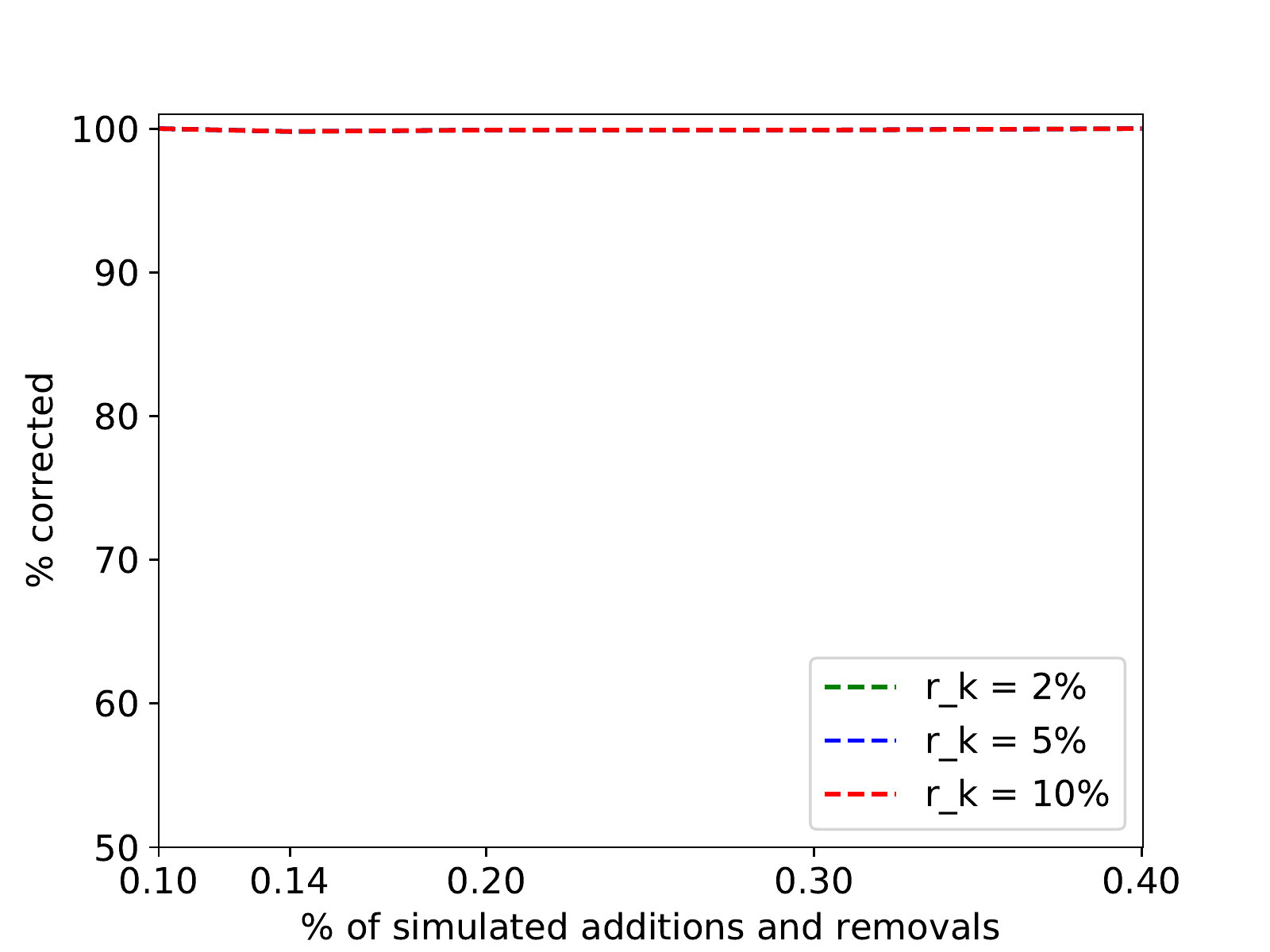} & 
  \includegraphics[width=0.5\columnwidth]{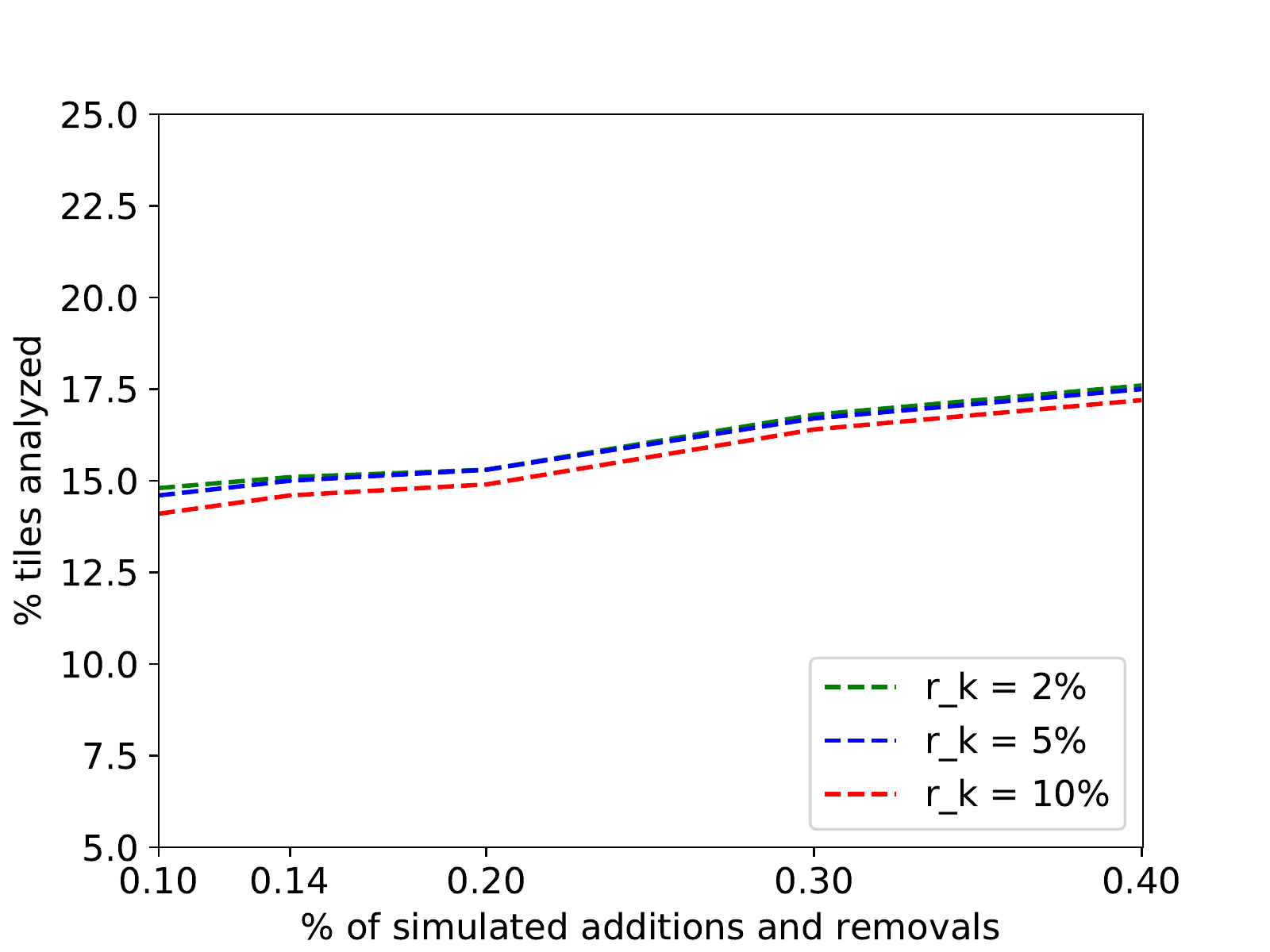} \\
  (a)  & (b)  \\  
\hline
\multicolumn{2}{c}{Zimbabwe dataset}\\
  \includegraphics[width=0.5\columnwidth]{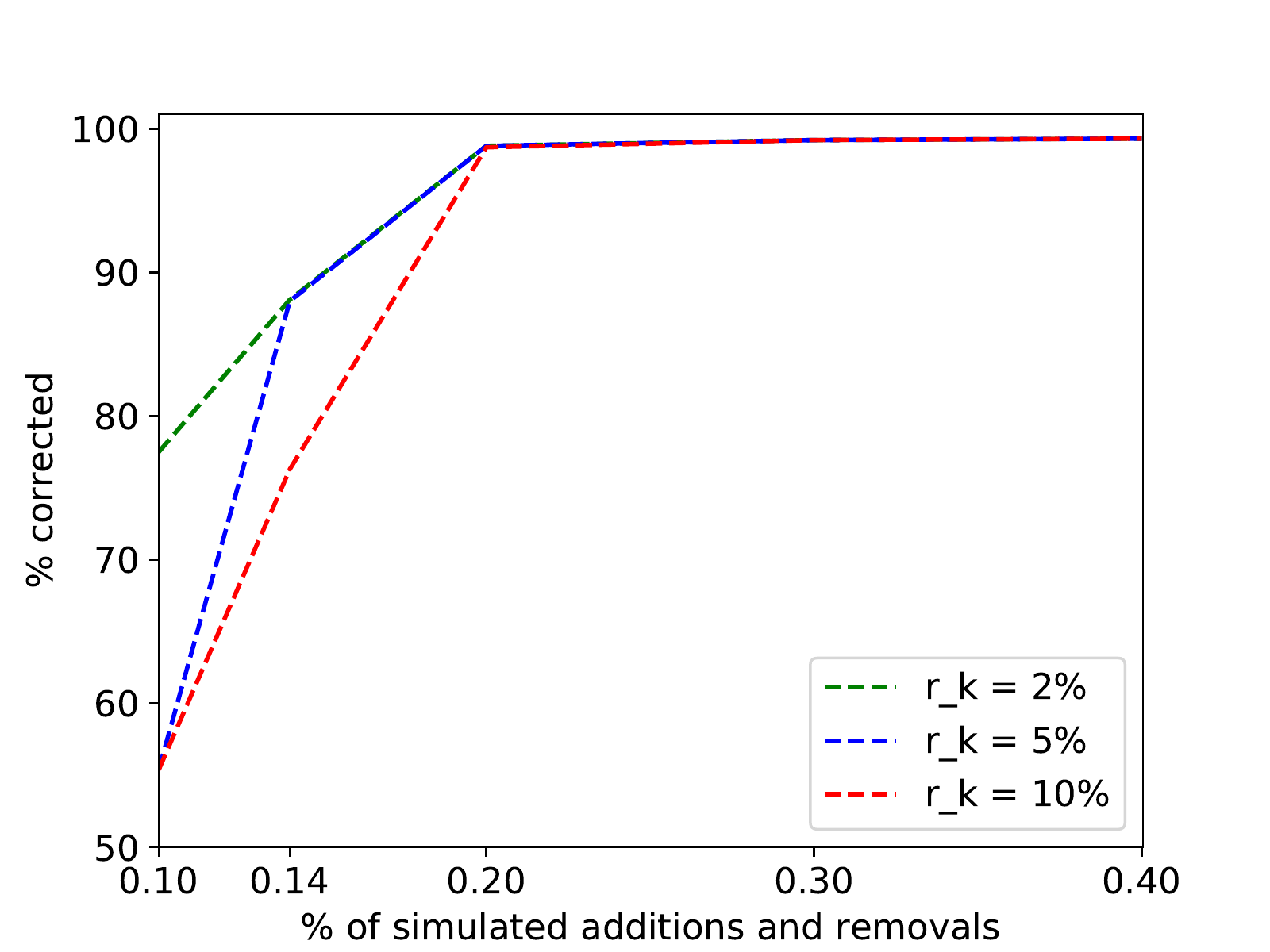} &
  \includegraphics[width=0.5\columnwidth]{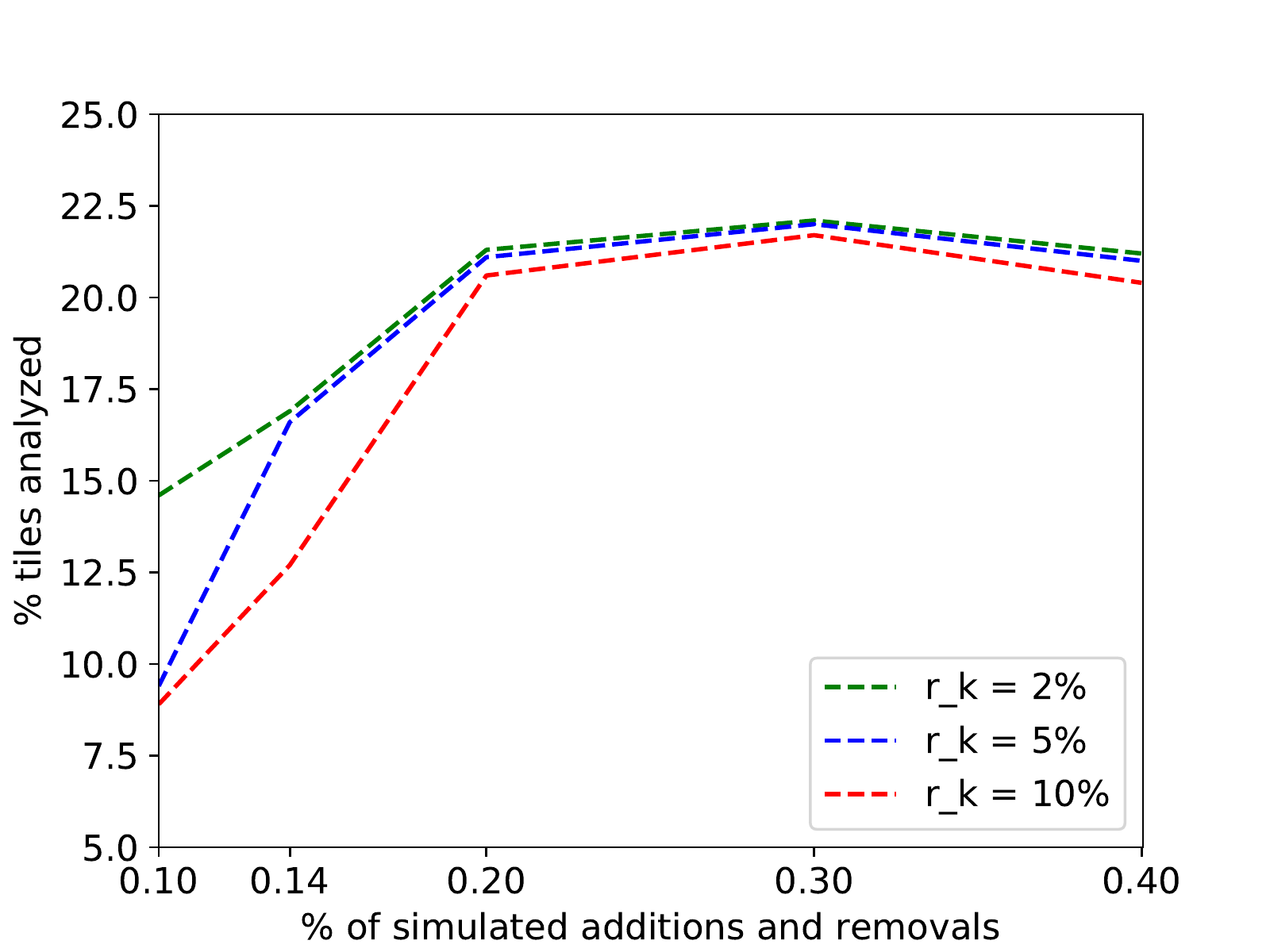} \\ 
  (c)  & (d) \\
\end{tabular} 
\end{center}
\caption{Stopping criteria performance using simulated addition, removals and small shifts. 
	\label{fig:simulation_experiments}}
\end{figure}

Figure~\ref{fig:simulation_experiments} shows the results of our proposed method with different stopping criteria parameter values on the Tanzania and Zimbabwe datasets with different quality levels of simulated wrong annotations (different percentages of simulated additions and removals). In the Tanzania dataset (see Figures~\ref{fig:simulation_experiments}a-b)), we can observe that even in the case of very low quality annotations our method can find almost all the wrong annotations for the three differents $r_k$ values (stopping criteria). This at the cost of analyzing around 14-17\% of the tiles in the Tanzania dataset. 

In the Zimbabwe dataset (see Figures~\ref{fig:simulation_experiments}c-d)), we can observe that the percentage of wrong annotations corrected varies with the quality level of the annotations and the $r_k$ values. When the quality of the annotations is reasonably good (10\% of simulated additions and removals) the annotation process stops too early because the probability maps are not of good quality in this dataset and it is more difficult to find tiles with wrong annotations. Thus, the percentage of wrong annotations corrected does not attain very high values. When the annotations are of bad quality (e.g., 40\% of simulated additions and removals) it is easier to find the tiles with wrong annotations and the percentage of wrong annotations corrected is very high, at the cost of analyzing around 20\% of the tiles in the Zimbabwe dataset. 

In general, when the probability maps are accurate higher values of $r_k$ can be used and therefore less amount of tiles will be analyzed by user annotators. On the other hand, when the probability maps are not reliable lower values of $r_k$ should be used to find most of the tiles with wrong annotations at the cost of correcting/verifying more tiles.

\subsection{Experiments with real users}
\label{subsec:real_user_experiments}

Table~\ref{tab:experiment_with_real_users} presents the results showing the percentage of wrong annotations corrected and percentages of tiles analyzed for different values of $r_k$ (stopping criteria parameter) in the two analyzed datasets. For the Mugumu dataset, we can observe that almost all the wrong annotations were corrected when the annotation process stops using the three values of $r_k$. The results in the Gweru are similar, but considerably more samples are analyzed when $r_k = 0.02$ than when $r_k = 0.10$. These results using real user annotation corrections confirm the effectiveness of the proposed method in a more realistic scenario.  During the experiment, the volunteers commented that at the beginning of the experiment most of the tiles require some annotation correction and the need for correction decreased over time.

\begin{table}[!t]
\centering

\caption{Results of the experiments with real user interactions. We report the percentage of wrong annotations corrected (\% corrected) and percentages of tiles analyzed for three different values of $r_k$ (stopping criteria parameter) in the Mugumu and Gweru datasets.}
\label{tab:experiment_with_real_users}
\begin{tabular}{|l|r|r|r|r|}
\hline
\multicolumn{5}{|c|}{Experiments with real user interactions}\\
\hline
$r_k$ (\%) &\multicolumn{2}{c|}{Mugumu dataset (Tanzania)} &\multicolumn{2}{c|}{Gweru dataset (Zimbabwe)} \\
\cline{2-5}
 &\multicolumn{1}{c|}{\% corrected}&\multicolumn{1}{c|}{\% tiles analyzed}  
 &\multicolumn{1}{c|}{\% corrected}&\multicolumn{1}{c|}{\% tiles analyzed} \\
\hline
10 & 99.0 & 22.7 & 97.9 & 23.7 \\ 
5  & 99.0 & 23.6 & 98.6 & 24.8 \\ 
2  & 99.0 & 24.6 & 98.6 & 28.7 \\ 
\hline
\end{tabular}

\end{table}

\section{Conclusions}
\label{sec:conclusions}

In this work, we proposed a methodology for the interactive correction/verification of rural building annotations in OpenStreetMap, which could be useful for the validation and quality improvement of OSM data.
The proposed methodology aims to quickly find a small number of regions that need to be verified by a user, avoiding the task of exhaustively verifying the large imagery obtained from the analyzed geographical area. In order to analyze such large images, we proposed an efficient CNN building segmentation method to obtain a building probability map. We evaluated several strategies to measure how wrong the current annotations are, based on both the current annotations and the generated building probability map. 
The sum of absolute differences between the original annotations and the probability map leads to better performances, among the evaluated tile selection strategies.
We also observed that the approach of retraining the CNN building segmentation model with the newly annotated/verified data considerably improves the accuracy of the method. The experiments that involve real user annotations show that the proposed stopping criterion allows our method to analyze less than a quarter of the total number of tiles obtained from the image and correct more than 98\% of the annotation errors. We believe that the proposed methodology could be very helpful for improving the annotation efficiency of humanitarian mapping projects, such as the Humanitarian OpenStreetMap Team~\footnote{https://www.hotosm.org/}. The proposed methodology not only makes more efficient the process of annotation but also produces a better experience for the volunteers, keeping high the levels of engagement (i.e., avoiding the analysis of uninteresting regions). As future work, we will propose a model that can estimate the number of edits needed to correct a group of annotations and also analyze the annotation skill of different users.

\section*{Acknowledgment}
The authors would like to thank Bing maps and OpenStreetMap for the access to the imagery and geographical objects' footprints respectively through their APIs.
This research was funded by Funda\c{c}\~{a}o de Amparo \`{a} Pesquisa do
Estado de S\~{a}o Paulo (FAPESP, grant 2016/14760-5 and 2014/12236-1), Conselho Nacional de Desenvolvimento Cient\'{\i}fico e Tecnol\'{o}gico (CNPq, grant 303808/2018-7), 
Coordena\c{c}\~{a}o de Aperfei\c{c}oamento de Pessoal de N\'{\i}vel Superior - Brasil (CAPES, finance code 001) and by the Swiss National Science Foundation (grant PP00P2-150593).

\section*{Data and codes availability statement}
The data and codes that support the findings of this study are available in figshare.com with the identifier(s) at the link \textbf{https://figshare.com/s/1147e0068fd9af8b1412}. Bing imagery cannot be made publicly available because of Bing's imagery data policies.

\bibliographystyle{tfv}
\bibliography{refs}

\end{document}